\documentclass[letterpaper]{article} 
\usepackage{aaai25}  
\usepackage{times}  
\usepackage{helvet}  
\usepackage{courier}  
\usepackage[hyphens]{url}  
\usepackage{graphicx} 
\usepackage{natbib}  
\usepackage{caption} 
\usepackage{algorithm}
\usepackage{booktabs}
\usepackage{algpseudocode}
\usepackage{csquotes}
\usepackage{graphicx}
\usepackage{amsmath}
\usepackage{pifont}
\usepackage{amssymb}
\usepackage{booktabs}
\usepackage{multicol}
\usepackage{multirow}
\usepackage{graphicx}
\usepackage{csquotes}
\usepackage{algpseudocode}
\usepackage{subfigure}
\usepackage{amssymb}
\usepackage{amsthm}
\usepackage{amsmath} 
\usepackage{amsfonts}
\usepackage{pifont}
\usepackage{subfigure}
\usepackage{xcolor}
\usepackage{array}
\usepackage{siunitx}
\usepackage{numprint}

\usepackage{newfloat}
\usepackage{listings}
\DeclareCaptionStyle{ruled}{labelfont=normalfont,labelsep=colon,strut=off} 
\floatstyle{ruled}
\newfloat{listing}{tb}{lst}{}
\floatname{listing}{Listing}
\title{UFO: Enhancing Diffusion-Based Video Generation with a Uniform Frame Organizer}
\author{
    Delong Liu\textsuperscript{\rm 1},
    Zhaohui Hou\textsuperscript{\rm 2},
    Mingjie Zhan\textsuperscript{\rm 2},
    Shihao Han\textsuperscript{\rm 2},
    Zhicheng Zhao\textsuperscript{\rm 1,3,}\thanks{Corresponding author},
    Fei Su\textsuperscript{\rm 1,3}
}
\affiliations{
    \textsuperscript{\rm 1}Beijing University of Posts and Telecommunications\\
    \textsuperscript{\rm 2}SenseTime\\
    \textsuperscript{\rm 3}Beijing Key Laboratory of Network System and Network Culture\\
    \{liudelong, zhaozc, sufei\}@bupt.edu.cn, \{houzhaohui, zhanmingjie, hanshihao\}@sensetime.com

}

\usepackage{bibentry}
\begin{document}
\maketitle

\begin{abstract}
Recently, diffusion-based video generation models have achieved significant success. However, existing models often suffer from issues like weak consistency and declining image quality over time. To overcome these challenges, inspired by aesthetic principles, we propose a non-invasive plug-in called Uniform Frame Organizer (UFO), which is compatible with any diffusion-based video generation model. The UFO comprises a series of adaptive adapters with adjustable intensities, which can significantly enhance the consistency between the foreground and background of videos and improve image quality without altering the original model parameters when integrated. The training for UFO is simple, efficient, requires minimal resources, and supports stylized training. Its modular design allows for the combination of multiple UFOs, enabling the customization of personalized video generation models. Furthermore, the UFO also supports direct transferability across different models of the same specification without the need for specific retraining. The experimental results indicate that UFO effectively enhances video generation quality and demonstrates its superiority in public video generation benchmarks. The code will be publicly available at \url{https://github.com/Delong-liu-bupt/UFO}.

\end{abstract}

\begin{figure*}[!t]
    \centering
    \includegraphics[width=0.96\linewidth]{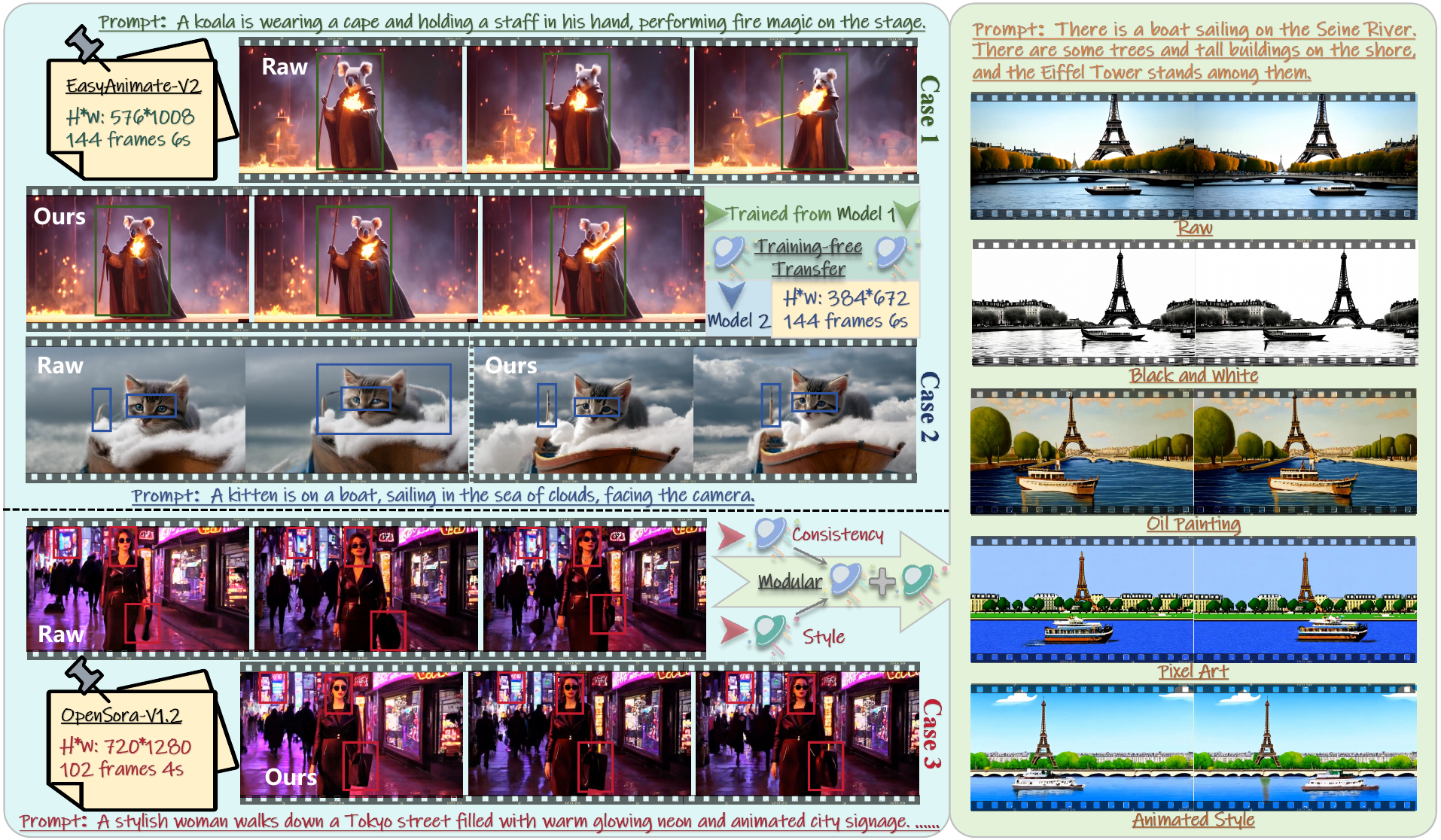}
    \label{fig1}
    \caption{The left side displays three cases: Cases 1 and 3 illustrate that our proposed consistency UFO can be integrated with the model to significantly \textbf{enhance the consistency} of generation. Case 2 demonstrates that the UFO can be \textbf{directly transferred} and effectively deployed between models of the same specification without the need for training. The cases on the right show that different consistency and stylization UFOs can be \textbf{freely combined} to customize video generators.}
\end{figure*}

\section{Introduction}
The rapid advancement of artificial intelligence has transformed the field of creative content generation. Individuals can quickly obtain personalized text \cite{text}, images \cite{image}, sounds\cite{sound}, and videos \cite{video} through simple natural language descriptions. In visual generation, diffusion models \cite{dm1,dm2}, which have excelled in image creation, play a crucial role. However, when applied to video generation, these models encounter challenges such as poor image quality, low aesthetic appeal, and weak consistency. For instance, as shown in Figure 1, even the most advanced open-source models cannot prevent subjects from changing shape throughout a video (e.g., the koala with the staff in Case 1, the kitten in Case 2, and the person with the bag in Case 3), or background inconsistencies (e.g., the boat in Case 2 and the advertising billboard in Case 3).

Aesthetic theory \cite{ugc1,ugc2} in visual media emphasizes the crucial roles of the consistency and clarity in enhancing viewer engagement and perceived quality. In video generation, where dynamic elements and transitions are essential, inconsistencies and blurring not only reduce aesthetic appeal but also undermine the effectiveness of visual communication. To address the challenges mentioned above, we propose the Uniform Frame Organizer (UFO), a non-invasive plug-in designed to enhance the consistency between the foreground and background and alleviate blurring issues, thereby improving video generation quality. Applicable to any diffusion-based video generation model, the UFO integrates a set of non-invasive adapters into the video generation model's backbone network, occupying only $0.005\times$ the size of the original model’s trainable parameters. These adapters are capable of autonomously adjusting their intensity of use, featuring a tunable intensity parameter, which is tuned to optimize the balance between dynamic visual content and static precision, reflecting a direct application of aesthetic principles in video generation.

Specifically, when using a small amount of video frames or images as training data, UFO sets the intensity to the highest value, dynamically controlling each adapter’s parameters and release intensity to force the model's output to approximate a static video, a scenario of extreme consistency. During this process, the UFO learns to identify and correct inconsistencies in videos. As the pre-trained model's parameters remain unchanged, the UFO's intensity can be adjusted to a lower value during application. This adjustment allows the model output to closely resemble the original while significantly enhancing the consistency between the subjects and the background in the video frames. It also markedly reduces issues such as sudden blurring of video frames.

To achieve the aesthetic consistency, during the training process, the primary optimization goal for the model integrated with the UFO is set to generate static video frames. This simplicity allows the model to learn quickly and converge after only 3000 training steps on a single GPU, using much fewer resources than fine-tuning or retraining video generation models. Moreover, once the parameters of the UFO are obtained, it supports direct transferability across multiple models of the same specification without the need for model-specific retuning (as shown in Figure 1, Case 2). Beyond enhancing video consistency, the UFO is capable of learning style variations from a limited amount of video-text pairs in the same style. It can also be combined flexibly with the consistency UFO to further enhance the production of videos that not only maintain consistency but also adhere more closely to specific stylistic preferences, as illustrated on the right side of Figure 1.

In practical applications, even the most advanced video generation models often require users to repeatedly adjust parameters and select results that meet their specific needs. In this process, some outcomes may become unusable due to minor consistency flaws or blurriness. The UFO resolves these issues without altering the original video content, significantly easing the challenge of achieving high-quality results. Practical tests on public video generation benchmarks Vbench \cite{vbench} demonstrate that the UFO notably enhances video consistency and quality. In summary, our main contributions are:

\begin{itemize}
    \item We propose the Uniform Frame Organizer (UFO), a non-invasive plug-in that obviously enhances video consistency and quality, and is compatible with any diffusion-based model. It features a novel adjustable intensity parameter for tuning of video effects.
    \item The UFO allows for direct transfer between models of the same specification and supports the modular integration of various UFOs, enabling the customization of personalized video generation models.
    \item Training UFO is very inexpensive, and enhances consistency without the need for video-text pairs.
    \item UFO significantly reduces the effort required by users to obtain high-quality videos, and the extensive experiments verify its efficiency and effectiveness.
\end{itemize}

\begin{figure*}[!t]
    \centering
    \includegraphics[width=0.95\linewidth]{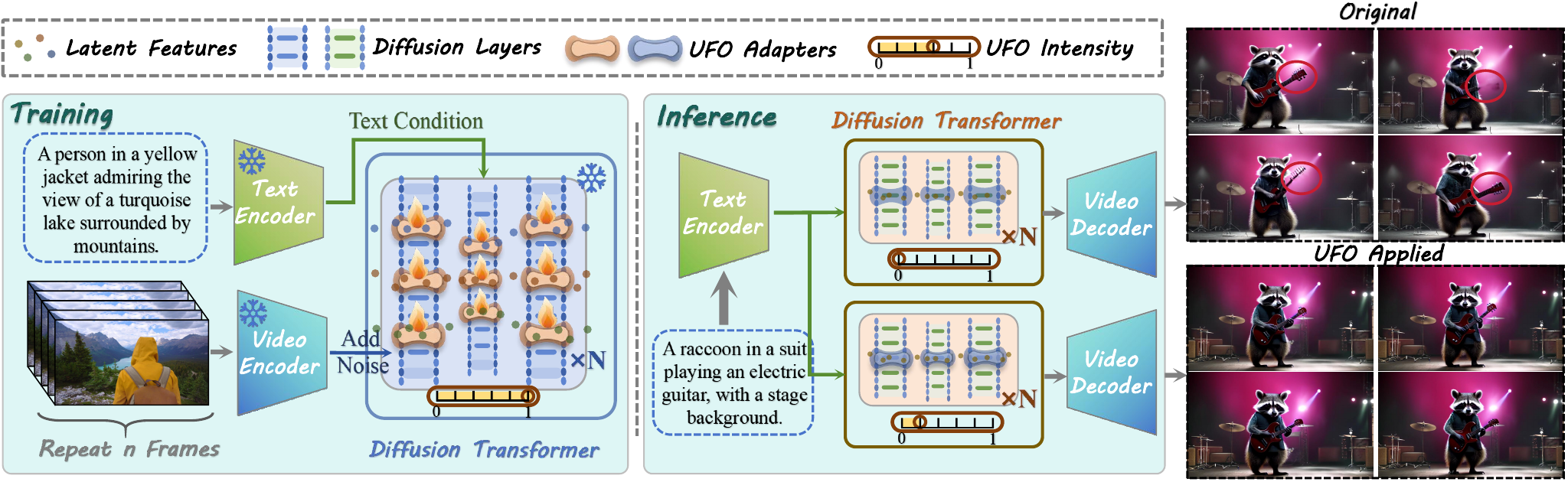}
    \label{fig2}
    \caption{Training and inference of consistency UFO. During training, all parameters of the original model are frozen, and the UFO operates at maximum intensity using image-text pair data, with images duplicated across multiple frames to meet training requirements. In inference, zero intensity mirrors the original generator, while low intensity improves video consistency. The right images compare these two scenarios.}
\end{figure*}

\section{Related Work}
The Diffusion Model (DM) has consistently excelled in image \cite{dmimg1,dmimg2,dmimg3} and video generation \cite{dmv1,dmv2,dmv3}, and has also expanded across various video generation tasks, including text-to-video \cite{t2v1,t2v2}, image-to-video \cite{i2v1,i2v2}, video-to-video \cite{v2v1,v2v2}, and applications under diverse control conditions such as pose \cite{pose1,pose2}, depth \cite{depth1,depth2}, and sketch \cite{ske1,ske2}. In the past two years, the text-to-video generation, our primary focus, has made rapid progress. Early work like Image Video \cite{imagenv} highlighted diffusion models' ability to produce high-quality videos. However, aligning videos precisely with text prompts while maintaining visual appeal remained challenging. Subsequent models, such as VideoCrafter \cite{vc1,vc2} and ModelScopeT2V \cite{mst2v}, were trained with large video datasets including WebVid-10M \cite{web} and InternVid \cite{intern}. These models enhanced the aesthetics of videos but often produced shorter videos with inconsistent quality and blurred visual details. The recent Gen-2 \cite{gen2} and Pika \cite{pika}, with their massive data and complex iterative sampling processes, enhanced video image quality. However, their stability and realism still need improvement, and the computational cost is also quite high.

The introduction of the Transformer-based Diffusion (DiT) architecture and it scalable parameter capability has opened new opportunities. Sora \cite{sora}, using DiT in its backbone network and exploiting vast datasets, has pioneered unparalleled zero-sample video generation, producing longer videos of higher quality. Although this technology is not yet publicly available, it has significantly spurred the open-source community's exploration of DiT-based video generation models \cite{open, open-plan, easy}, leading to a series of high-quality video generation models that substantially surpass previous models. Nonetheless, as video length increases, challenges like video consistency and blurring persist. In response, we propose a low-cost, widely applicable consistency enhancement plug-in, UFO, which is validated on the best-performing open-source models \cite{open, easy}.

\section{Methodology}
As shown in Figure 2, the UFO includes a series of lightweight adapters (Section 3.1) that can be non-destructively attached to any mapping layer of the model without altering the original model's parameters. During training (Section 3.2), only the UFO's parameters are updated, while the intensity is set to the highest to achieve extremely consistent videos under text conditions, rendering all video frames static. This phase drives the UFO to develop the ability to identify and correct inconsistencies. During inference (Section 3.3), setting the UFO's intensity to 0 will result in an output that is identical to that of the original pre-trained model. When it is at low level, the output will closely resemble the original, maintaining motion in video frames. The UFO's targeted repair capabilities enhance the consistency between subjects and backgrounds and mitigate video quality degradation. For example, when applying the same prompt and fixed random seed, the UFO-generated appearance and attire of the raccoon are more consistent, notably preventing significant shape transformations in the electric guitar being played, as shown in Figure 2.

\subsection{Lightweight Adapters for UFO}
To achieve cost-effective improvements in video generation models  and eliminate reliance on a single model framework, inspired by efficient parameter fine-tuning methods \cite{adapter, lora}, we design a series of adapters, each of which consists of a layer with minimal input or output dimensions, and is injected into the diffusion model with minimal overhead. These adapters act as the smallest sub-units for controlling the consistency of hidden features in video frames, enabling precise, targeted consistency corrections.

Specifically, in a module parameterized by \( \textbf{W} \in \mathbb{R}^{m \times n} \) in the DIT, we learn a detection layer \( \textbf{v}_{det} \in \mathbb{R}^{n \times d} \) to precisely locate features affecting video consistency. Concurrently, a correction layer \( \textbf{v}_{cor} \in \mathbb{R}^{m \times d} \) modifies the identified features. Here, \(d\) is chosen to be small to ensure parameter efficiency. Consequently, the original representation \( \textbf{y} = \textbf{W}\textbf{x} \) is modified as follows:
\[ \textbf{y} = \textbf{W}\textbf{x} + \alpha \beta (\textbf{v}_{det}^T \textbf{x}) \cdot \textbf{v}_{cor} \]
where \( \textbf{x} \in \mathbb{R}^n \) and \( \textbf{y} \in \mathbb{R}^m \) represent the input and output of the intermediate layer, respectively, the superscript \( T \) indicates transposition, \( \alpha \) is a adjustable intensity factor, and \( \beta \) is a learnable dynamic intensity factor, aiming to dynamically adjust the strength of each adapter of the UFO to ultimately correct the consistency of video frames.

Focusing on just two frames \(y_t, y_{t+n} \in \textbf{y}\) in the video, the difference in output between these two intermediate layers \( \Delta y_n = y_t - y_{t+n} \) can be expressed as:
\begin{align*}
\Delta y_n &= (\textbf{W}x_t - \textbf{W}x_{t+n}) + \alpha \beta ((\textbf{v}_{det}^T x_t) \cdot \textbf{v}_{cor} - (\textbf{v}_{det}^T x_{t+n}) \cdot \textbf{v}_{cor}) \\
&= \textbf{W}\Delta x_n + \alpha \beta \Delta (\textbf{v}_{det}^T x \cdot \textbf{v}_{cor}).
\end{align*}
During the training phase, with \( \alpha=1 \), the target is for all video frames to be identical, thus \( \Delta y_n=0 \) regardless of the value of \(n\). Therefore, the optimization goal for each latent feature is \( -\textbf{W}\Delta x_n = \beta \Delta (\textbf{v}_{det}^T x \cdot \textbf{v}_{cor})\), meaning that the trained \( \beta \), \( \textbf{v}_{det} \), and \( \textbf{v}_{cor} \) can adaptively identify and fill the variations in each video frame.

Utilizing this feature, during inference, \( \alpha \) is set to a low value, ensuring that the variability \( \Delta y_n \approx \textbf{W}\Delta x_n \) in video frames maintains the subjects, background, and motion capabilities essentially consistent with those of the pre-trained model's output videos, while the additional term \( \alpha \beta \Delta (\textbf{v}_{det}^T x \cdot \textbf{v}_{cor}) \) has a comprehensive view of the changes in video frames, adaptively enhancing the consistency of the output video. Furthermore, if the intensity factor \( \alpha \) is fixed, UFO, due to its parametric characteristics, also supports using a small batch of video-text pairs to directionally fine-tune the video generation model, customizing the video generation effects.

\subsection{Training of UFO}

During the training phase, video data \( \textbf{V} \in \mathbb{R}^{F \times H \times W \times C}\) is first compressed into a latent space representation \( z = \mathcal{E}(\textbf{V}) \) using a pretrained variational autoencoder (VAE) \cite{vae}. Additionally, a textual condition \( c \) is introduced, which is derived from a text encoder using prompts aligned with the video content. In the generation process, the diffusion model gradually introduces noise to simulate the diffusion of video data, forming perturbed samples \( z_t = \sqrt{\overline{\alpha}_t} z + \sqrt{1 - \overline{\alpha}_t} \epsilon\), where \( \epsilon \sim N(0, 1) \) represents noise sampled from a standard normal distribution, and \( \overline{\alpha}_t \) serves as a noise scheduler, with \( t \) denoting the diffusion time step.

After integration with UFO, the parameters of the original model are denoted as \( \theta \), and only the parameters within UFO are updated during training. The reverse diffusion process, which is essentially training the model to denoise, aims to predict the less noisy \( z_{t-1} \): \( p_\theta(z_{t-1} | z_t) = N(\mu_\theta(z_t), \Sigma_\theta(z_t)) \). Here, the log likelihood of the variational lower bound simplifies to \( L_{vlb}(\theta) = - \log p(z_0 | z_1, c) + \sum_t D_{KL} \left(q(z_{t-1} | z_t, z_0) \| p_\theta(z_{t-1} | z_t)\right) \). Since both \( q \) and \( p_\theta \) are Gaussian, the \( D_{KL} \) term is determined by the mean \( \mu_\theta \) and covariance \( \Sigma_\theta \). The \( \mu_\theta \) is reparametrized into the denoising model \( \epsilon_\theta \), which can be trained using a simple objective:
\[
L_{simple}(\theta) = \mathbb{E}_{z \sim p(z), \epsilon \sim N(0,1), t, c} \left[ \| \epsilon - \epsilon_\theta(z_t, t, c) \|_2^2 \right],
\]
According to \cite{vlb}, it is necessary to fully optimize the \( D_{KL} \) term (i.e., train using the full \( L_{vlb } \)) to train an LDM with learnable covariance \( \Sigma_\theta \). Therefore, the training loss for UFOs employs both \( L_{simple} \) and \( L_{vlb} \).

When training the consistency UFO, every frame in \( \textbf{V} \) used is identical, thus image-text pairs, which are more readily available, can be used as training data. For customizing stylization UFOs, regular video-text pairs are used as training data.

\subsection{Inference}

During inference, the trained UFO is integrated into the diffusion model, retaining all functionalities of the original model. For the consistency UFO, its intensity factor \( \alpha \) tends to be set to a low value, which can be adjusted based on the performance of the original pre-trained model during video inference. If issues such as inconsistency or blurring are severe, \( \alpha \) should be increased, which enhances video frame consistency. This adjustment allows users to control video consistency according to their needs. For stylization UFOs, \( \alpha \) is suggested to match the level used during training, and minor adjustments can optimize personalization. Note that when combining different UFOs, the intensity of each UFO needs to be adjusted as required.

\begin{table*}[!t]
\centering
\resizebox{\linewidth}{!}{
\begin{tabular}{@{}c|c|c|c|ccccc|ccc|c|c@{}}
\toprule
\textbf{No.} & \textbf{Model} & \textbf{Resolution} & \textbf{UFO/\( \alpha \)} & \textbf{TQ} & \textbf{SC} & \textbf{BC} & \textbf{TF} & \textbf{MS} & \textbf{FWQ} & \textbf{AQ} & \textbf{IQ} & \textbf{SQ} & \textbf{EC} \\ 
\midrule
1 & \multirow{9}{*}{Open} & \multirow{3}{*}{$240 \times 426$} & 0 & 95.55\% & 93.00\% & 94.91\% & 97.84\% & 96.43\% & 55.25\% & 51.96\% & 58.54\% & 66.68\% & - \\
2 &  &  & 0.1 & 96.46\% & 94.53\% & 95.52\% & 98.49\% & 97.29\% & 55.43\% & 52.25\% & 58.61\% & 67.15\% & 51/1165 \\
3 &  &  & 0.2 & \textbf{97.01\%} & \textbf{95.08\%} & \textbf{96.32\%} & \textbf{98.71\%} & \textbf{97.91\%} & \textbf{56.23\%} & \textbf{53.05\%} & \textbf{59.40\%} & \textbf{67.78\%} & 86/1165 \\ 
 \cline{3-14}
4 &  & \multirow{3}{*}{$480 \times 854$} & 0 & 95.24\% & 93.04\% & 93.62\% & 98.30\% & 95.99\% & 59.49\% & 57.01\% & 61.96\% & 72.13\% & - \\
5 &  &  & 0.1 & 96.35\% & 94.45\% & 94.88\% & 98.97\% & 97.08\% & 60.19\% & 57.63\% & 62.75\% & 72.24\% & 35/1165 \\
6 &  &  & 0.2 & \textbf{97.00\%} & \textbf{95.48\%} & \textbf{95.68\%} & \textbf{99.14\%} & \textbf{97.70\%} & \textbf{60.50\%} & \textbf{58.03\%} & \textbf{62.96\%} & \textbf{72.40\%} & 80/1165 \\ 
 \cline{3-14}
7 &  & \multirow{3}{*}{$720 \times 1280$} & 0 & 95.33\% & 92.47\% & 95.22\% & 98.38\% & 95.23\% & 60.42\% & 57.27\% & 63.57\% & 73.00\% & - \\
8 &  &  & 0.1 & 96.70\% & 94.50\% & 96.55\% & 99.06\% & 96.67\% & 60.61\% & 57.59\% & 63.62\% & 73.78\% & 39/1165 \\
9 &  &  & 0.2 & \textbf{97.26\%} & \textbf{95.34\%} & \textbf{97.08\%} & \textbf{99.27\%} & \textbf{97.36\%} & \textbf{60.97\%} & \textbf{57.91\%} & \textbf{64.03\%} & \textbf{73.80\%} & 71/1165 \\ 
\hline
10 & \multirow{6}{*}{Easy} & \multirow{3}{*}{$384 \times 672$} & 0 & 97.07\% & 94.67\% & 96.84\% & 99.49\% & 97.26\% & 63.53\% & 62.78\% & 64.27\% & 71.21\% & - \\
11 &  &  & 0.07 & 98.20\% & 96.44\% & 97.73\% & 99.68\% & 98.94\% & 63.64\% & 62.95\% & 64.32\% & 71.48\% & 28/1165 \\
12 &  &  & 0.15 & \textbf{99.02\%} & \textbf{98.06\%} & \textbf{98.54\%} & \textbf{99.81\%} & \textbf{99.66\%} & \textbf{63.77\%} & \textbf{63.05\%} & \textbf{64.48\%} & \textbf{71.82\%} & 81/1165 \\ 
\cline{3-14}
13 &  & \multirow{3}{*}{$576 \times 1008$} & 0 & 96.41\% & 92.94\% & 96.49\% & 99.30\% & 96.91\% & 63.57\% & 61.43\% & 65.71\% & 70.58\% & - \\
14 &  &  & 0.07 & 97.53\% & 94.80\% & 97.52\% & 99.57\% & 98.22\% & 64.42\% & 62.08\% & 66.75\% & 71.64\% & 21/1165 \\
15 &  &  & 0.15 & \textbf{98.57\%} & \textbf{96.69\%} & \textbf{98.63\%} & \textbf{99.68\%} & \textbf{99.28\%} & \textbf{64.98\%} & \textbf{62.31\%} & \textbf{67.65\%} & \textbf{72.08\%} & 75/1165 \\
\bottomrule
\end{tabular}
}
\caption{Impact of the consistency UFO on the performance of different base models. The values for TQ and FWQ represent the mean scores across their respective dimensions.}
\end{table*}

\begin{figure*}[!t]
    \centering
    \includegraphics[width=0.95\linewidth]{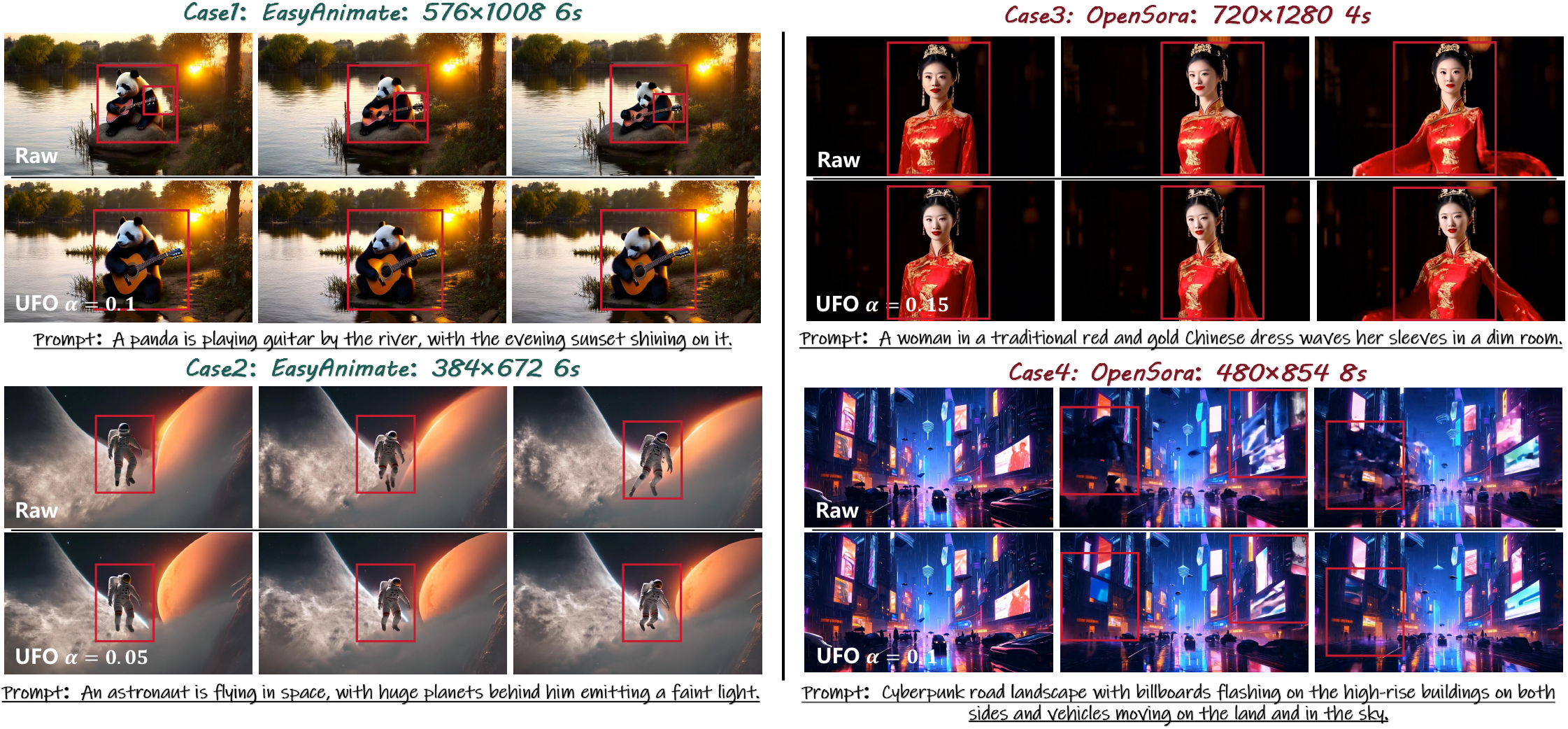}
    \label{fig3}
    \caption{Visualizations of the consistency UFO. The areas highlighted in red boxes show inconsistencies or blurriness in the videos produced by the pre-trained model.}
    
\end{figure*}

\begin{figure*}[!t]
    \centering
    \includegraphics[width=0.95\linewidth]{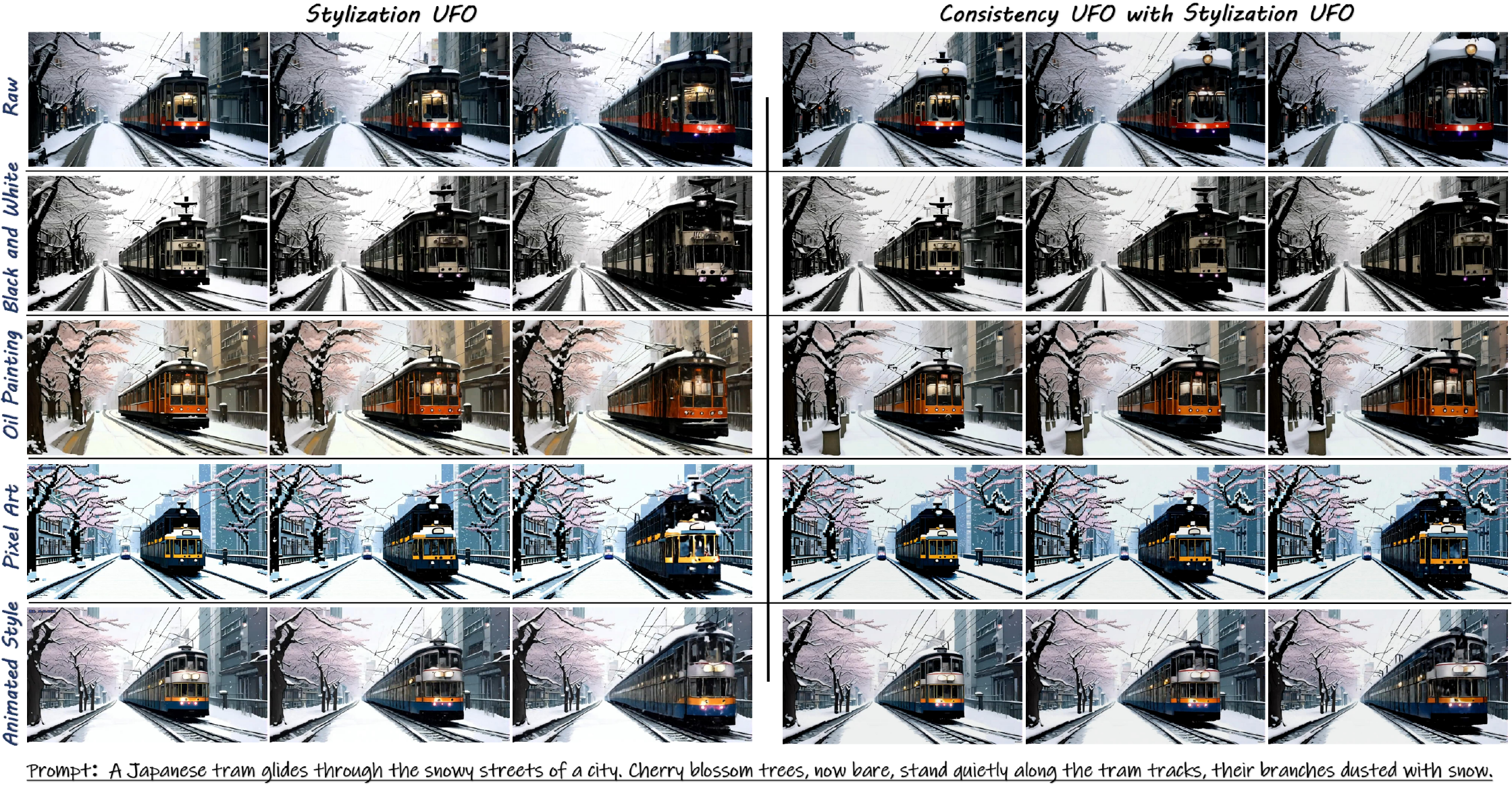}
    \caption{Examples of the effects of consistency UFO with stylization UFO. The first row illustrates the results without using stylization UFO, while rows two to five demonstrate the effects of different stylization UFOs. Videos on the right have added consistency UFO compared to those on the left. In these cases, all the stylization UFOs have \( \alpha=1 \) and the consistency UFO have \( \alpha=0.1 \), all generated by Open with a resolution of $720 \times 1280$ and a duration of 4 seconds.}
    
\end{figure*}

\section{Experimental Results}
\subsection{Settings}
\textbf{Implementation details.} To ensure the rigor of our experiments, we train UFOs using two of the latest text-to-video open-source models, EasyAnimate-V2 (Easy) \cite{easy} and OpenSora-V1.2 (Open) \cite{open}. Training is conducted on 4 NVIDIA A100 GPUs, with inference running on a single GPU. During the training process, only the parameters of the UFOs are updated, with each UFO undergoing 3000 training steps. All adapters have a hyperparameter dimension $d=4$, and gradient accumulation is not used. For Open, a linear warm-up strategy is employed in the first 500 steps, where the learning rate gradually increases from nearly zero to $2e-4$, and this rate is maintained after the warm-up phase. For Easy, the learning rate is set at $1e-4$ and remains constant. The rest of the training settings follow the original methods. During inference, all settings use the recommended configurations of the original methods, with videos set at 24 Frames Per Second (FPS), and all experiments and visual effects in the text use the same random seed to compare with and without UFOs. More details on training and inference can be found in the supplementary materials.

\textbf{Training Datasets.} For training the consistency UFO, we use a subset of the LAION-Aesthetics V2 \cite{laion} dataset with aesthetic scores above 6.5, from which we extract 12K image-text pairs to create static video-text pairs for training. For the training of stylization UFOs, we collect 300 videos for each of the four styles (Pixel Art, oil painting, animated style, black and white) from publicly available video resources on the internet. The text for these videos is automatically annotated using the 13B version of the PLLaVA \cite{pllava} model, with descriptions regarding the video style removed during training.

\textbf{Evaluation Metrics.} To objectively demonstrate the improvements in video consistency and quality achieved by the UFO, we employ the latest video generation evaluation method, Vbench, using a fixed intensity setting for the consistency UFO. This evaluation encompasses two main dimensions: Video Quality (VQ) and Semantic Quality (SQ). As our approach does not specifically target enhancements in video semantic consistency, our primary focus is on the VQ metrics. These include four dimensions of ``Temporal Quality" (TQ): ``Subject Consistency" (SC), ``Background Consistency" (BC), ``Temporal Flickering" (TF), ``Motion Smoothness" (MS), and two dimensions of ``Frame-Wise Quality" (FWQ): ``Aesthetic Quality" (AQ) and ``Imaging Quality" (IQ). We also assess the dimensions related to SQ, providing only a total score. For a single complete evaluation, a total of 4720 videos inferred from Vbench’s official prompts are used, of which 1165 videos relate to the four dimensions of TQ. Since some pre-trained model inferences produce videos with minimal visual changes, using a fixed intensity UFO can cause the frames to become nearly static, potentially skewing the TQ metrics. Consequently, we exclude such videos from the evaluation, recorded as ``Excluded Count" (EC), to reflect the impact of the consistency UFO. More details on the criteria for judging near-static conditions and the specifics of the metrics are available in the supplementary materials.

\subsection{Quantitative Results.}
We evaluate the consistency UFO on two baseline models, Easy and Open. Open supports high-quality video generation across multiple resolutions with a single model, while Easy performs poorly when handling different resolutions, requiring the use of two separate models for videos of various resolutions. Videos used for Vbench evaluation are rendered at typical resolutions supported by the original models, with each video running for 4 seconds. The results are shown in Table 1. When \( \alpha = 0 \), it reflects the performance without UFO, while \( \alpha > 0 \) indicates the use of UFO at varying intensities. In our primary dimension of concern, Temporal Quality (TQ), it is clear that UFO significantly enhances both the consistency between the subject and background, and the smoothness of video motion. A higher intensity of UFO leads to more pronounced improvements, but it may also cause more videos with minimal dynamics to become static. However, in practical use, users can freely adjust the intensity of UFO based on video outcomes, thus avoiding such issues. Similarly, the Frame-Wise Quality (FWQ) dimension related to image quality shows the same trend because UFO effectively eliminates blurring and flickering issues in the video, thereby enhancing image quality. Surprisingly, UFO also results in gains in the Semantic Quality (SQ) dimension, likely due to enhancements in TQ and FWQ dimensions that improve the visual expression stability of the generated videos.

Notably, although Easy uses two different models of the same specification to process videos of two resolutions, the same consistency UFO plugin is utilized. It was only trained on the model handling higher resolutions, suggesting that UFO can effectively transfer between models and achieve the desired effects.

\subsection{Qualitative Results}
\textbf{Consistency UFO}
In Figure 3, we showcase four qualitative results on two baseline models to illustrate the intuitive effects and characteristics of our consistency UFO. In Case 1, without the use of the consistency UFO, the appearance and size of the panda, as well as the guitar, are inconsistent. After applying the UFO with \( \alpha = 0.1 \), the panda and the guitar maintain consistency, preserving the composition and main elements of the original video. Case 2 demonstrates that transferring the consistency UFO directly to another model is also effective, significantly improving the consistency of the astronaut depicted in the image. Case 3 displays the universality of the UFO, which is effective under any framework, significantly enhancing the appearance and posture consistency of the woman in the image. Case 4 primarily shows that the consistency UFO can address issues of blurring and flickering in long video generation, effectively alleviating the blurring of billboards and the flickering of black objects in long videos. In practical applications, the intensity level can be adjusted based on the degree of inconsistency or blurriness in the original video to optimize the output.

\textbf{Consistency UFO with Stylization UFO}
Due to the parametric characteristics of UFO, various styles of stylization UFOs can be customized. Figure 4 demonstrates the effects of combining a pre-trained model with stylization UFOs, which can transform original videos into various styles while preserving the fundamental elements and layout of the original image. For example, despite the videos on the left side of the figure having different styles, elements such as the direction of the train, the cherry trees on the left side of the road, the buildings on the right, and the shooting angle remain consistent. However, this transformation still preserves the subject inconsistencies present in the original video. Therefore, by leveraging the flexibility of UFO, combining stylization UFO with consistency UFO can produce more consistent personalized videos. Comparing the video frames on the left and right sides of the figure, it is noticeable that maintaining consistency in the tram’s front and doors is challenging across all styles, with even the colors (oil painting style) and proportions (animated style) varying. These issues are effectively addressed in the video frames on the right.

\begin{table}[!t]
\centering
\resizebox{\linewidth}{!}{
\begin{tabular}{@{}l|c|c|c|c|c|c|c@{}}
\toprule
\textbf{Model} & \textbf{Params.} & \textbf{Time} & \textbf{UFO/\( \alpha \)} & \textbf{\(d\)} & \textbf{TQ} & \textbf{FWQ} & \textbf{SQ} \\ 
\midrule
\multirow{6}{*}{Open} & 1.14 B & -  & 0 & - & 95.24\% & 59.49\% & 72.13\% \\
 & 1.42 M & 0.24\% & 0.1 & 1 & 95.37\% & 59.43\% & 72.09\% \\
 & 2.83 M & 0.33\% & 0.1 & 2 & 95.98\% & 59.89\% & 72.19\% \\
 & 5.66 M & 0.40\%  & 0.1 & \textbf{4} & \textbf{96.35\%} & \textbf{60.19\%} & 72.24\% \\
 & 11.32 M & 0.51\% & 0.1 & 8 & 96.29\% & 60.11\% & \textbf{72.38\%} \\
 & 90.56 M & 1.08\% & 0.1 & 64 & 96.03\% & 59.67\% & 71.98\% \\
\midrule
\multirow{6}{*}{Easy} & 818.17 M & - & 0 & - & 96.41\% & 63.57\% & 70.58\% \\
 & 1.89 M & 0.31\% & 0.07 & 1 & 96.53\% & 63.76\% & 70.76\% \\
 & 3.79 M & 0.40\% & 0.07 & 2 & 97.16\% & 64.28\% & 71.12\% \\
 & 7.58 M & 0.49\% & 0.07 & \textbf{4} & \textbf{97.53\%} & 64.42\% & \textbf{71.64\%} \\
 & 15.15 M & 0.58\% & 0.07 & 8 & 97.49\% & \textbf{64.50\%} & 71.62\% \\
 & 121.21 M & 1.44\% & 0.07 & 64 & 97.14\% & 63.97\% & 70.89\% \\
\bottomrule
\end{tabular}
}
\caption{Performance changes associated with different dimensions \(d\) in the consistency UFO adapters. The `Params.' column represents the amount of the diffusion model's parameters when UFO is not used, and the trainable parameters when UFO is in use. `Time' indicates the percentage increase in time required to infer a single video compared to the original model. All performance metrics are based on the inference of 4-second videos, with a resolution of \(480 \times 854\) for Open and \(576 \times 1008\) for Easy.}
\end{table}

\subsection{Ablition studies.}

\begin{figure}[!t]
    \centering
    \subfigure{
        \includegraphics[width=0.49\linewidth]{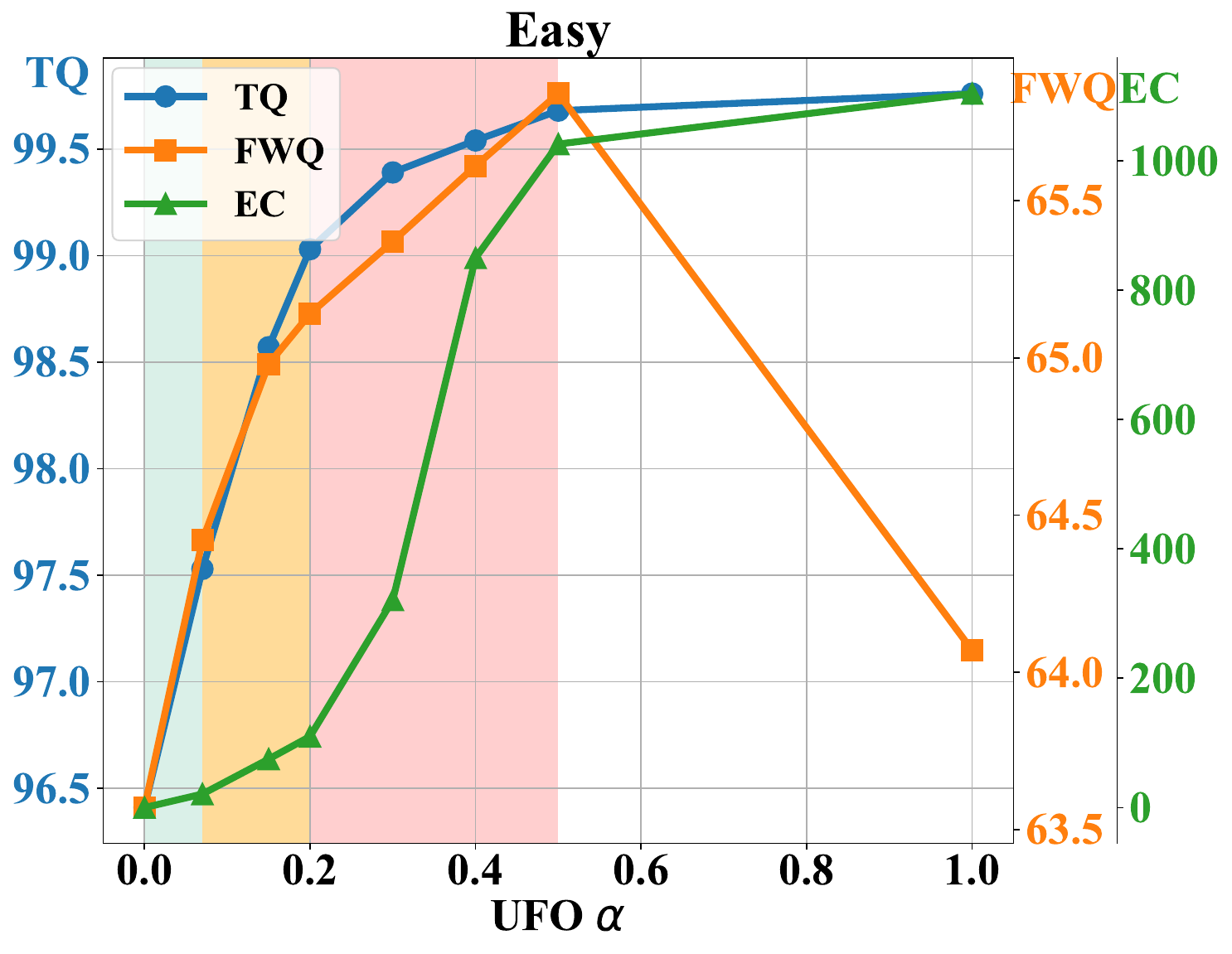}
        \label{fig:easy}
    }
    \hspace{-0.35cm} 
    \subfigure{
        \includegraphics[width=0.49\linewidth]{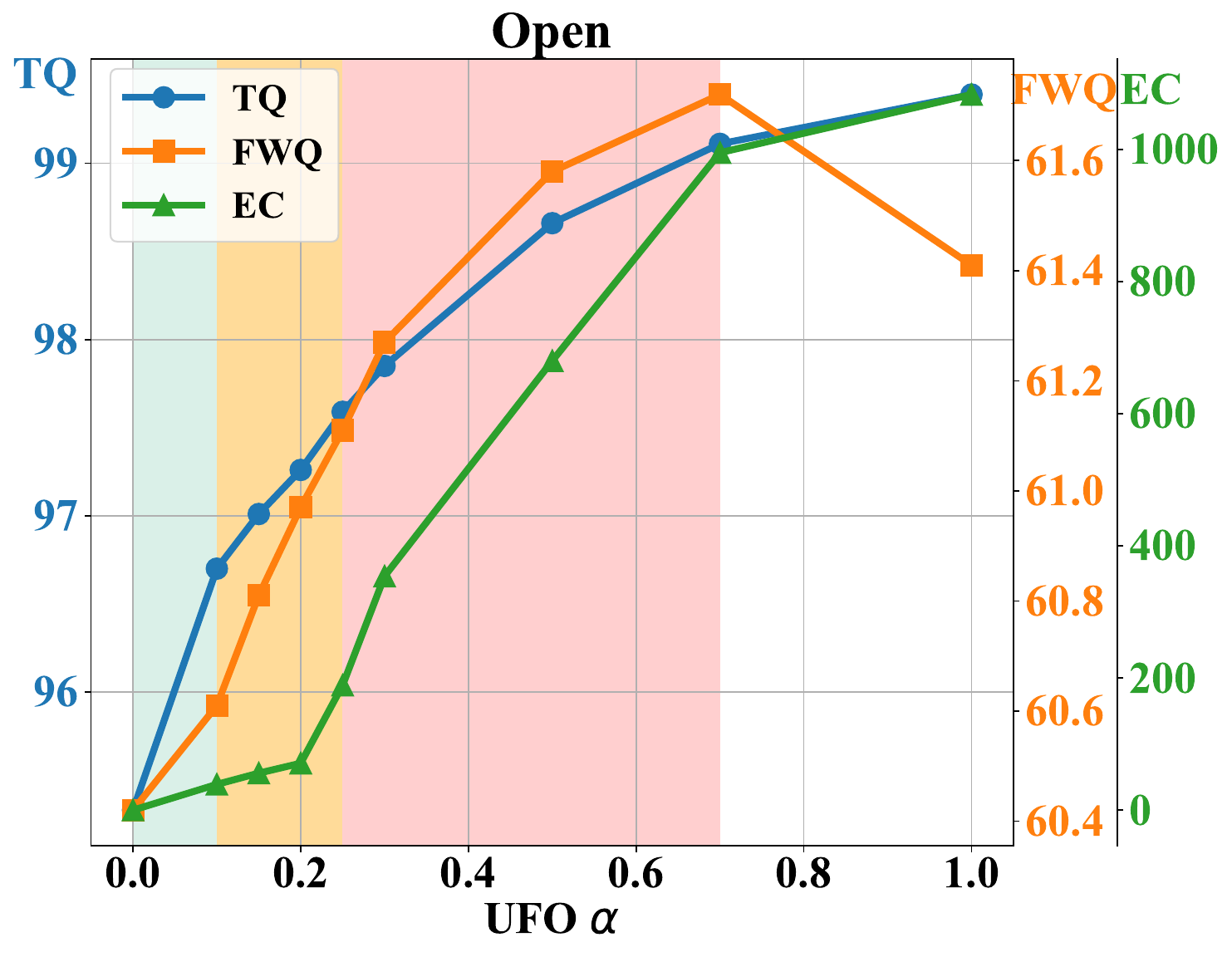}
        \label{fig:open}
    }
    \caption{Metric variations with different \( \alpha \) levels for the consistency UFO. Left image metrics are from a 4-second video at \(576 \times 1008\) on Easy. Right image metrics are from a 4-second video at \(720 \times 1280\) on Open. Light blue indicates conservative strategy, orange for moderate, and red for aggressive.}
    \label{fig:overall}
\end{figure}

\textbf{UFO Adapters Dimension.} Table 2 illustrates the performance changes when different dimensions \(d\) are used in the consistency UFO adapters. It is evident that increasing \(d\) up to 4 effectively enhances both consistency and image quality. However, further increases beyond 4 do not yield additional gains. When \(d\) becomes too large, the UFO begins to fit the characteristics of the limited data used for training. While this still can enhance the consistency of the images, it causes the content of the visuals to diverge from those generated by the original model. Although increasing \(d\) does not significantly reduce inference speed, it introduces more parameters, tailoring the model more closely to the specific characteristics of the training data. Therefore, after comprehensive consideration, \(d=4\) is established as the optimal setting for the UFO.

\textbf{UFO Intensity \( \alpha \).} To achieve optimal visual effects for specific prompts, one can initially obtain preliminary results from the video generation model and then adjust the consistency UFO's \( \alpha \) based on the degree of inconsistency observed. For scenarios requiring enhanced general generative capabilities, \( \alpha \) needs to be preset. Figure 5 illustrates the impact of the consistency UFO with fixed \( \alpha \) values, showing similar metric trends across two models. With a conservative strategy, videos maintain their dynamism while improving in consistency and quality. A moderate strategy enhances these aspects significantly, though it may slow down motion in some videos. An aggressive strategy markedly increases consistency and quality but can lead many videos to become nearly static. Further increasing \( \alpha \) risks making nearly all generated videos unusable. Since videos generated by the Easy model exhibit inherently less motion than those from Open, \( \alpha \) settings are adjusted more conservatively across these strategies.

\begin{figure}[!t]
    \centering
    \includegraphics[width=1\linewidth]{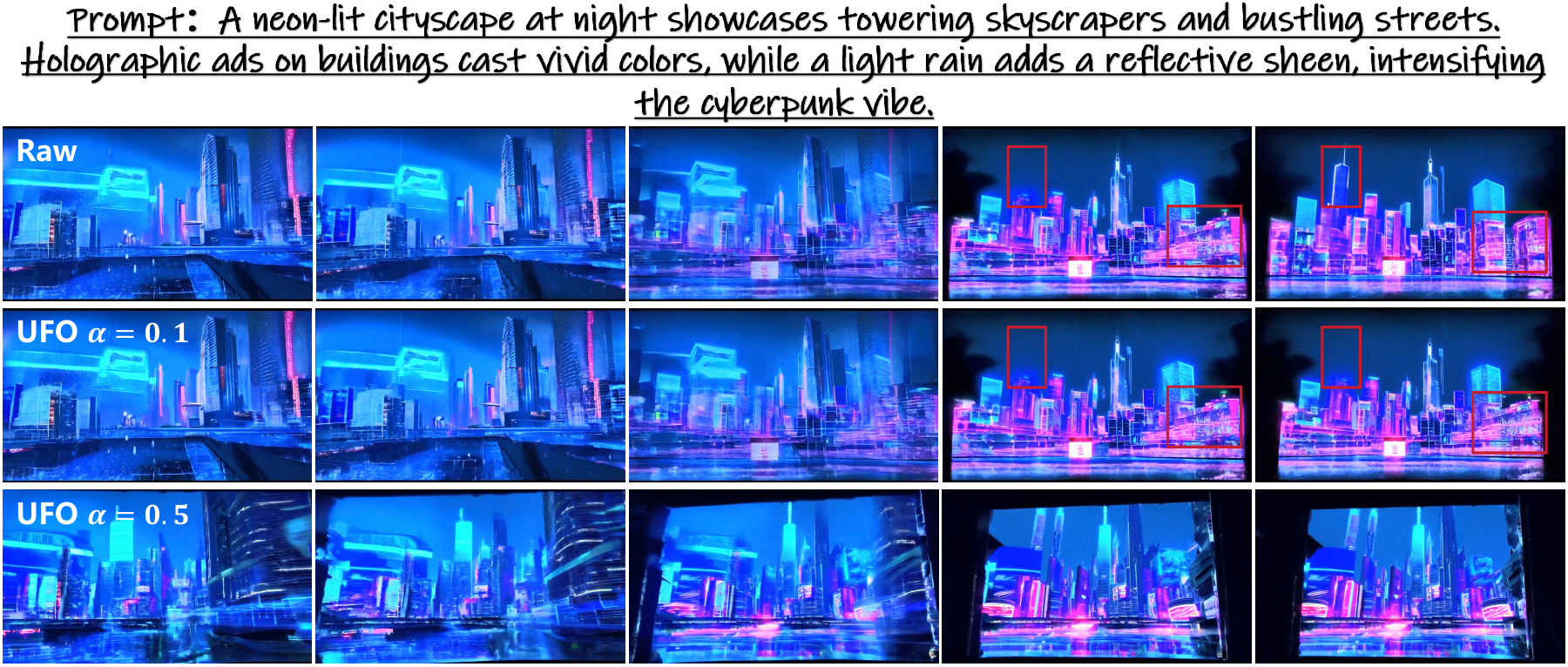}
    \caption{Special cases requiring high-intensity consistency UFO. The example in the image is from OpenSora-inferred 720P 8-second videos, with red boxes highlighting the inconsistencies in the original video.}
    
\end{figure}

\section{Special Case}
Although the UFO significantly enhances video consistency and quality, its design principle ensures that its outputs strictly adhere to the visuals produced by the pre-trained model. Consequently, the video generation capabilities of the UFO are limited by the underlying pre-trained model, and it struggles to address anomalies in videos such as unexpected scene transitions or extremely poor image quality. For instance, as shown in Figure 6, when a scene transition occurs, using the consistency UFO typically improves the image quality and consistency across both scenes, yet it fails to maintain a single unchanged scene effectively. In such cases, one might consider forcibly increasing the intensity of the UFO to facilitate smoother transitions, albeit at the cost of some fidelity, thereby rendering the video usable. However, any choice involves certain trade-offs; thus, a possible direction for future improvements in UFO is to maintain the main elements of the original output while achieving higher consistency in videos, even at higher intensities.

\section{Conclusion}
In this paper, we propose and validate the UFO, a non-invasive plugin for diffusion-based video generation models. By integrating the UFO into existing models, its effectiveness in mitigating common problems like video quality degradation and frame inconsistency is demonstrated, and without significantly increasing computational demands. In addition, The proposed intensity \( \alpha \) also provides users with the flexibility to control video consistency, facilitating the creation of videos that meet their specific needs. Moreover, the UFO's modular design and low resource requirements make it easily transferable between different models, thus enhancing their flexibility and scalability. In the future, we aim to improve the UFO so that it can reliably enhance video quality by automatically intensity adjusting.

\bibliography{aaai25}

\appendix
\clearpage
\section{Additional Details}
\subsection{Evaluation Metrics.}
\textbf{Vbench Metrics.}
In this study, we utilize the Vbench \cite{vbench} evaluation system for a comprehensive assessment of video quality. This system categorizes video quality metrics into two main domains: Temporal Quality (TQ) and Frame-Wise Quality (FWQ), each further divided into multiple specific indices to capture various aspects of video quality. Temporal Quality is essential for ensuring a consistent viewing experience throughout the video sequence. This category includes several components: Subject Consistency (SC), which gauges the consistency of subjects, such as people or objects, based on feature similarity between frames; Background Consistency (BC), assessing the stability of background scenes across frames; Temporal Flickering (TF), measured by calculating the average absolute difference between frames to spotlight inconsistencies in local and high-frequency details; and Motion Smoothness (MS), which ensures that video movements comply with the physical laws of the real world. Frame Quality evaluates the quality of each individual frame independently of its temporal context, focusing on Aesthetic Quality (AQ) and Imaging Quality (IQ). AQ appraises the artistic and visual appeal of each frame, considering factors such as layout, color coordination, and overall aesthetics. Conversely, IQ examines the technical aspects of each frame, including exposure levels, noise, and clarity.

Beyond these quantitative metrics, we also assess the semantic quality of generated videos using the Semantic Quality (SQ) component of the Vbench system, which includes two semantically-related dimensions: Semantics and Style. The Semantics dimension assesses whether the video content accurately portrays the entities and their attributes described in the text prompts, ensuring that the objects, actions, and colors in the video correspond with the descriptions. The Style dimension evaluates whether the visual style of the generated videos meets the specified user requirements, ensuring that the videos are not only compliant in content but also visually appealing and stylistically consistent.

\begin{figure}[!t]
    \centering
    \includegraphics[width=1\linewidth]{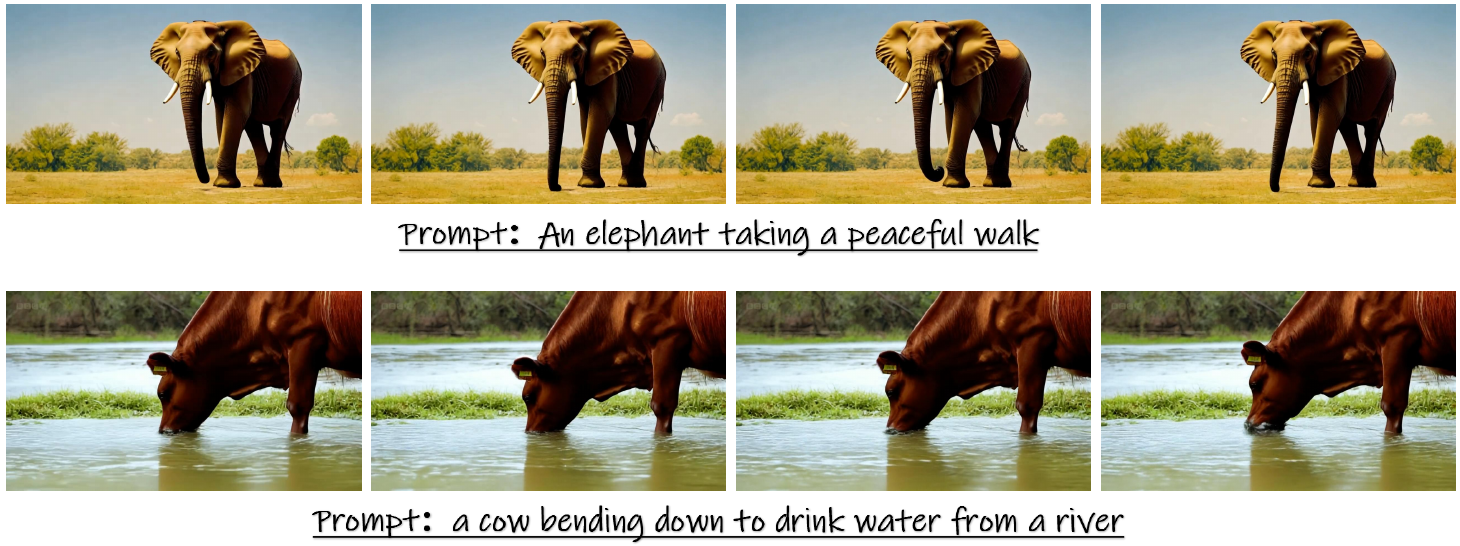}
    \caption{Examples of 4-second videos generated by the models with an \(OFT \approx 1\).}
    
\end{figure}

\textbf{Excluded Count Metric.}
To ensure a fairer assessment of TQ, we exclude videos that tend to become static after the addition of the consistency UFO. The number of videos removed is defined as the Excluded Count (EC), which measures the losses incurred due to using fixed intensities of UFO. Specifically, we use RAFT \cite{raft} to estimate the intensity of optical flow between consecutive frames of the generated videos. Considering that some videos only require movement of small objects, we take the average of the highest 5\% of the optical flow values as the basis for determining if a video is static, defined as the Optical Flow Threshold (OFT). EC counts the videos where the OFT value is less than 1 and has decreased by more than 1.5 times compared to the OFT value of the original video after using the consistency UFO. Figure 7 displays videos generated with an \(OFT \approx 1\), where the backgrounds tend to be static, and the subjects move more slowly compared to typical videos, thus selecting this value as the threshold.

\subsection{Training and Inference Details}

\textbf{Training Details.}
During training, both methods employe data bucketing techniques with varying resolutions and aspect ratios, aligning with the pre-training settings of the original models to cover a broad range of data. The T5 (Flan-T5-XXL) model \cite{t5} serves as the text encoder. For the consistency UFO, image data is duplicated to match the specific video lengths required for model training. In contrast, normal video-text pair data is used for training the stylization UFO.

\begin{figure*}[!t]
    \centering
    \includegraphics[width=0.98\linewidth]{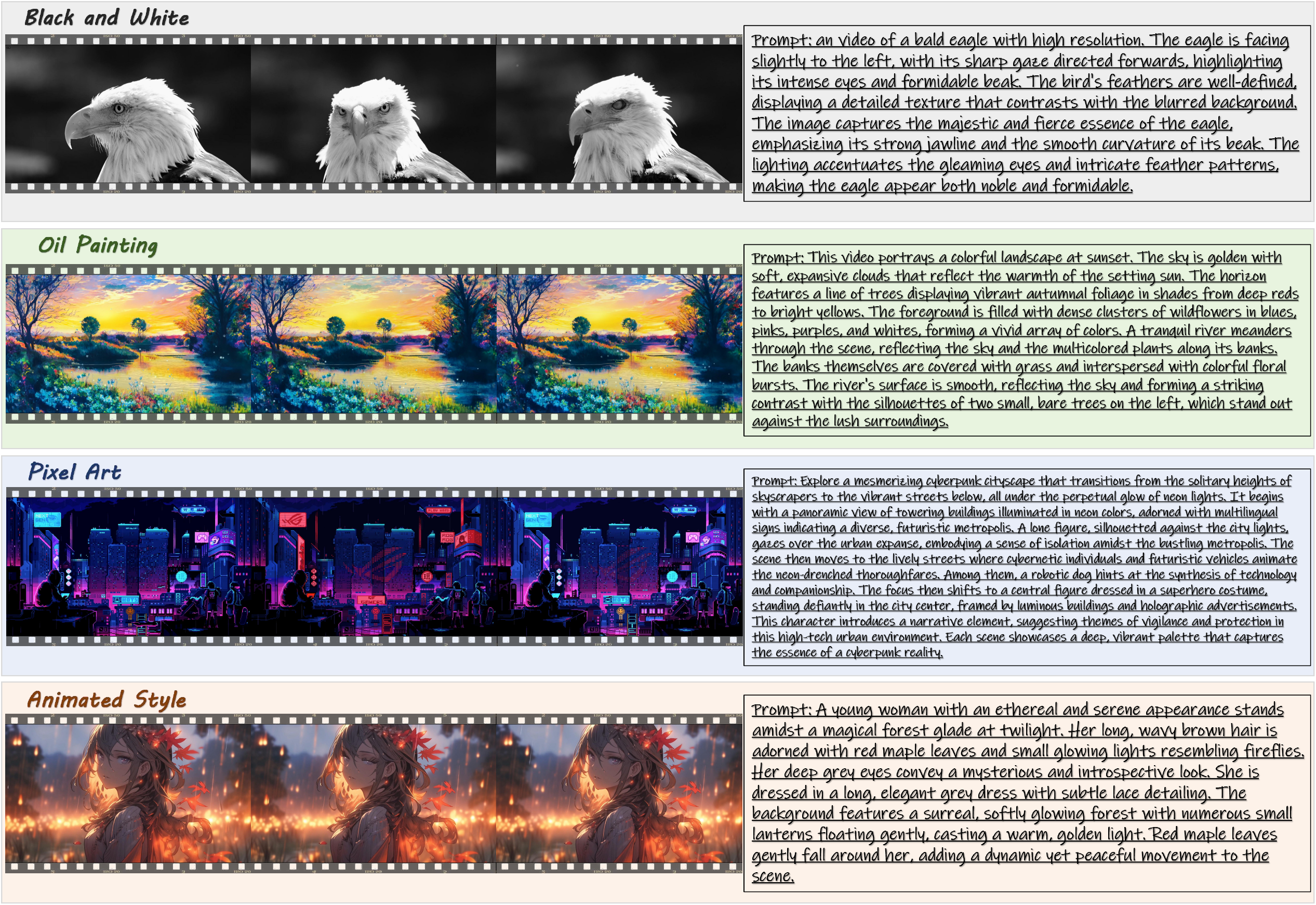}
    \caption{Examples of video-text pairs in four styles used for training the stylization UFO.}
    
\end{figure*}

Training for the stylization UFO utilizes video data sourced from the internet, which initially lacks any accompanying text descriptions. Therefore, we opt to use PLLaVA \cite{pllava} for automatic text annotation by uniformly extracting four frames from each video, following procedures from the open-source code repository \cite{open}. The automatically generated text may contain descriptions pertaining to the video's style. Since the stylization UFO is designed for a single style and should avoid reliance on specific prompts, we employ GPT-4 \cite{gpt4} to remove style-related terms from the text. The used prompt is: "Rewrite this prompt to exclude any descriptions of video styles such as cartoon, oil painting, black-and-white, pixel, etc. Focus on describing the content of the image. Output only the rewritten result without any additional output: {Caption}." Examples of the video-text pairs obtained are shown in Figure 8.

\textbf{Inference Details.} During inference, both models adhere to the official recommended settings. For OpenSoraV1.2 \cite{open}, the video output consists of 51 frames for every 2 seconds of video, and the model utilizes 30 sampling steps per video. For EasyAnimateV2 \cite{easy}, the output is 24 frames per second of video, and each video undergoes 50 sampling steps. EasyAnimateV2 also supports the use of negative prompts, applying a uniform negative prompt during video inference: ``The video is not of high quality, it has low resolution, and the audio quality is not clear. Strange motion trajectory, poor composition and deformed video, low resolution, duplicate and ugly, strange body structure, long and strange neck, bad teeth, bad eyes, bad limbs, bad hands, rotating camera, blurry camera, shaking camera. Deformation, low-resolution, blurry, ugly, distortion.”

\begin{table*}[!t]
\centering
\resizebox{\linewidth}{!}{
\begin{tabular}{@{}c|c|c|ccccc|ccc|c|c@{}}
\toprule
No. & Method & UFO/\( \alpha \) & TQ & SC & BC & TF & MS & FWQ & AQ & IQ & SQ & EC \\ 
\midrule
1 & Raw & 0 & 97.07\% & 94.67\% & 96.84\% & 99.49\% & 97.26\% & 63.53\% & 62.78\% & 64.27\% & 71.21\% & - \\ \midrule
2 & Transferred & \multirow{2}{*}{0.07} & \textbf{98.20\%} & 96.44\% & 97.73\% & \textbf{99.68\%} & \textbf{98.94\%} & 63.64\% & 62.95\% & 64.32\% & 71.48\% & 28/1165 \\
3 & Retrained &  & 98.18\% & \textbf{96.52\%} & \textbf{97.78\%} & 99.61\% & 98.79\% & \textbf{63.73\%} & \textbf{63.03\%} & \textbf{64.42}\% & \textbf{71.53\%} & 27/1165 \\  \midrule
4 & Transferred & \multirow{2}{*}{0.15} & \textbf{99.02\%} & 98.06\% & \textbf{98.54\%} & 99.81\% & \textbf{99.66\%} & 63.77\% & 63.05\% & 64.48\% & 71.82\% & 81/1165 \\
5 & Retrained &  & 99.00\% & \textbf{98.10\%} & 98.51\% & \textbf{99.83\%} & 99.57\% & \textbf{63.83}\% & \textbf{63.14\%} & \textbf{64.52\%} & \textbf{71.88\%} & 79/1165 \\
\bottomrule
\end{tabular}
}
\caption{Performance comparison of the consistency UFO under different strategies. In the `Method' column, `Raw' refers to using the original model, `Transferred' refers to using a consistency UFO trained on another model, and `Retrained' refers to retraining the consistency UFO on the current model. All performances are inferred from 4-second videos at a resolution of $384 \times 672$ on EasyAnimateV2.}
\end{table*}

\begin{table}[!t]
\centering
\resizebox{\linewidth}{!}{
\begin{tabular}{@{}c|l|c|c|c|c|c|c@{}}
\toprule
\textbf{Model} & \textbf{Resolution} & \textbf{UFO/\( \alpha \)} & \textbf{Durs} & \textbf{TQ} & \textbf{$\Delta$TQ} & \textbf{FWQ} & \textbf{$\Delta$FWQ} \\ 
\midrule
\multirow{8}{*}{Open} & \multirow{4}{*}{$480 \times 854$} & 0 & 4s & 95.24\% & - & 59.49\% & - \\
 & & 0.1 & 4s & \textbf{96.35\%} & +1.11\% & 60.19\% & +0.70\% \\
 & & 0 & 8s & 94.78\% & - & 59.79\% & - \\
 & & 0.1 & 8s & 96.03\% & \textbf{+1.25\%} & \textbf{60.56\%} & \textbf{+0.77\%} \\
\cline{2-8}
 & \multirow{4}{*}{$720 \times 1080$} & 0 & 4s & 95.33\% & - & 60.42\% & - \\
 & & 0.1 & 4s & \textbf{96.70\%} & +1.37\% & 60.61\% & +0.19\% \\
 & & 0 & 8s & 94.95\% & - & 60.60\% & - \\
 & & 0.1 & 8s & 96.43\% & \textbf{+1.48\%} & \textbf{61.10\%} & \textbf{+0.50\%} \\
\hline
\multirow{8}{*}{Easy} & \multirow{4}{*}{$384 \times 672$} & 0 & 4s & 97.07\% & - & 63.53\% & - \\
 & & 0.07 & 4s & \textbf{98.20\%} & +1.13\% & \textbf{63.64\%} & +0.11\% \\
 & & 0 & 6s & 96.57\% & - & 62.32\% & - \\
 & & 0.07 & 6s & 98.04\% & \textbf{+1.47\%} & 63.12\% & \textbf{+0.80\%} \\
\cline{2-8}
 & \multirow{4}{*}{$576 \times 1008$} & 0 & 4s & 96.41\% & - & 63.57\% & - \\
 & & 0.07 & 4s & \textbf{97.53\%} & +1.12\% & \textbf{64.42\%} & +0.85\% \\
 & & 0 & 6s & 96.24\% & - & 63.34\% & - \\
 & & 0.07 & 6s & 97.49\% & \textbf{+1.25\%} & 64.31\% & \textbf{+0.97\%} \\
\bottomrule
\end{tabular}
}
\caption{Performance for different cideo durations. ``Durs" refers to video lengths. The symbol \(\Delta\) represents the difference in each metric compared to the performance metrics when the UFO is not used.}
\end{table}

\subsection{Additional Results}
\textbf{Transferability of UFO.} In tests conducted with EasyAnimateV2, we utilize two different models to infer videos at varying resolutions, but only use one consistency UFO, demonstrating its training-free transfer capabilities. Table 3 delves further into this feature, showing that using a consistency UFO trained on same-architecture models on new models performs almost identically to retraining, from a performance perspective. These results further suggest that the consistency UFO has learned a higher-dimensional ability to detect and correct inconsistencies that is independent of the original model's parameters.

\textbf{Video Duration.} Table 4 displays how various metrics change when videos of different lengths are generated using the model combined with the consistency UFO. By examining TQ for videos of different durations created under the same settings with the same model, it is evident that the longer the video, the lower its temporal consistency. However, changes in \(\Delta\)TQ indicate that the benefits provided by using the UFO increase with video length. Similarly, the \(\Delta\)FWQ metric shows that the longer the video duration, the greater the improvements attributed to UFO.

\begin{figure*}[!t]
    \centering
    \includegraphics[width=0.98\linewidth]{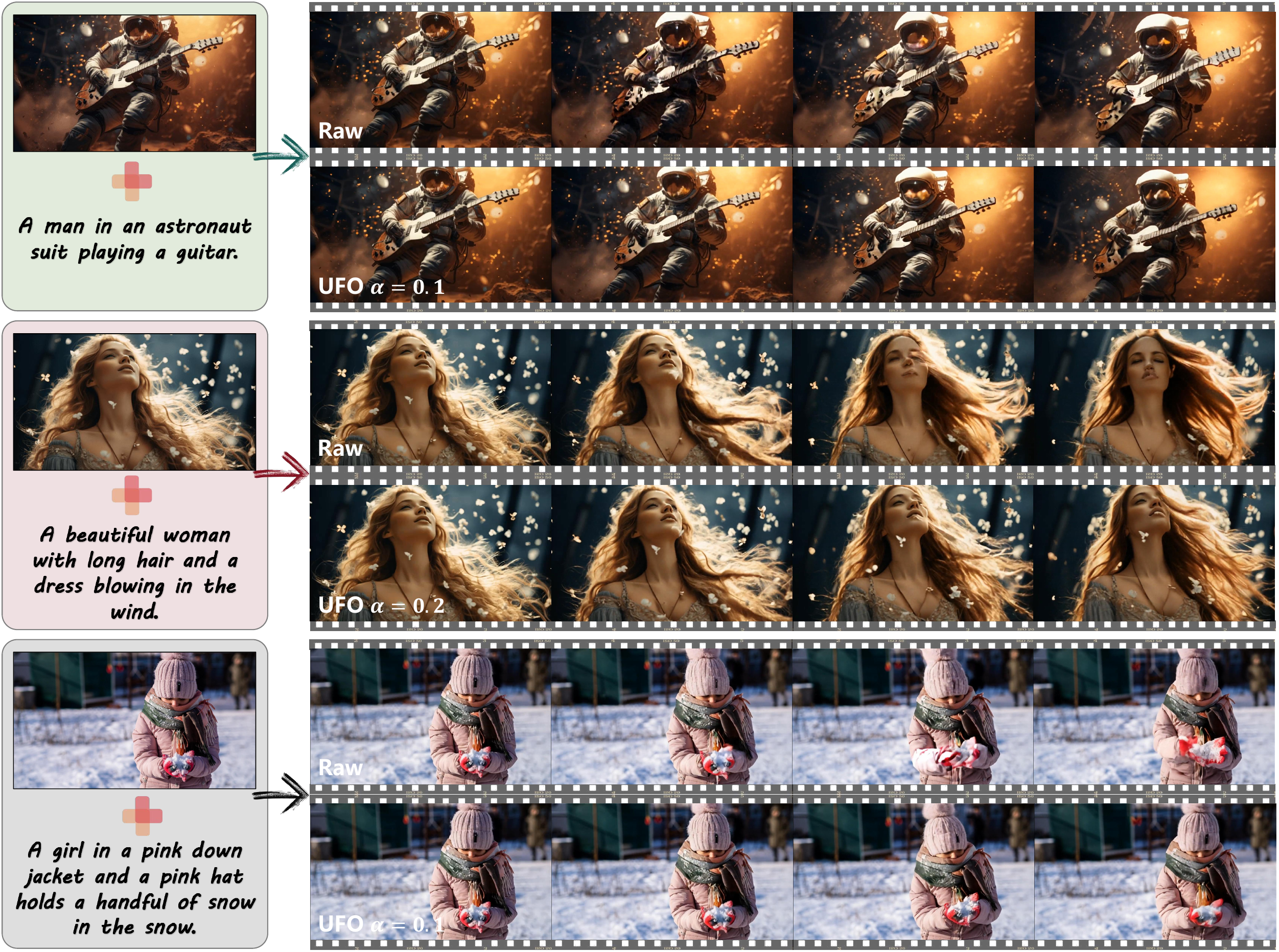}
    \caption{Visualization examples of the consistency UFO applied to video generation from images. All examples were generated by OpenSoraV1.2 under the combined influence of reference images and descriptive text, with a resolution of $720 \times 1280$ and a duration of 4 seconds.}
\end{figure*}

\section{Additional Visualizations}
\textbf{Visualization of the Consistency UFO Applied to Video Generation from Images.} Unlike EasyAnimateV2, OpenSoraV1.2 also supports the feature of using a reference image as the first frame of the generated video, which likewise faces consistency issues. Similarly, using the consistency UFO can significantly improve this inconsistency, with visualizations shown in Figure 9. After applying the UFO, a clear enhancement in video consistency is evident (for example, the astronaut's hands and guitar, the woman's face, and the little girl's hands and hat).

\textbf{More Cases for the Combination of Consistency UFO and Stylization UFO.} Figure 10 presents additional cases of freely combining different UFOs. From the images, it is evident that freely combining UFOs can generate videos that are both more consistent and personalized.

\begin{figure*}[!t]
    \centering
    \includegraphics[width=1\linewidth]{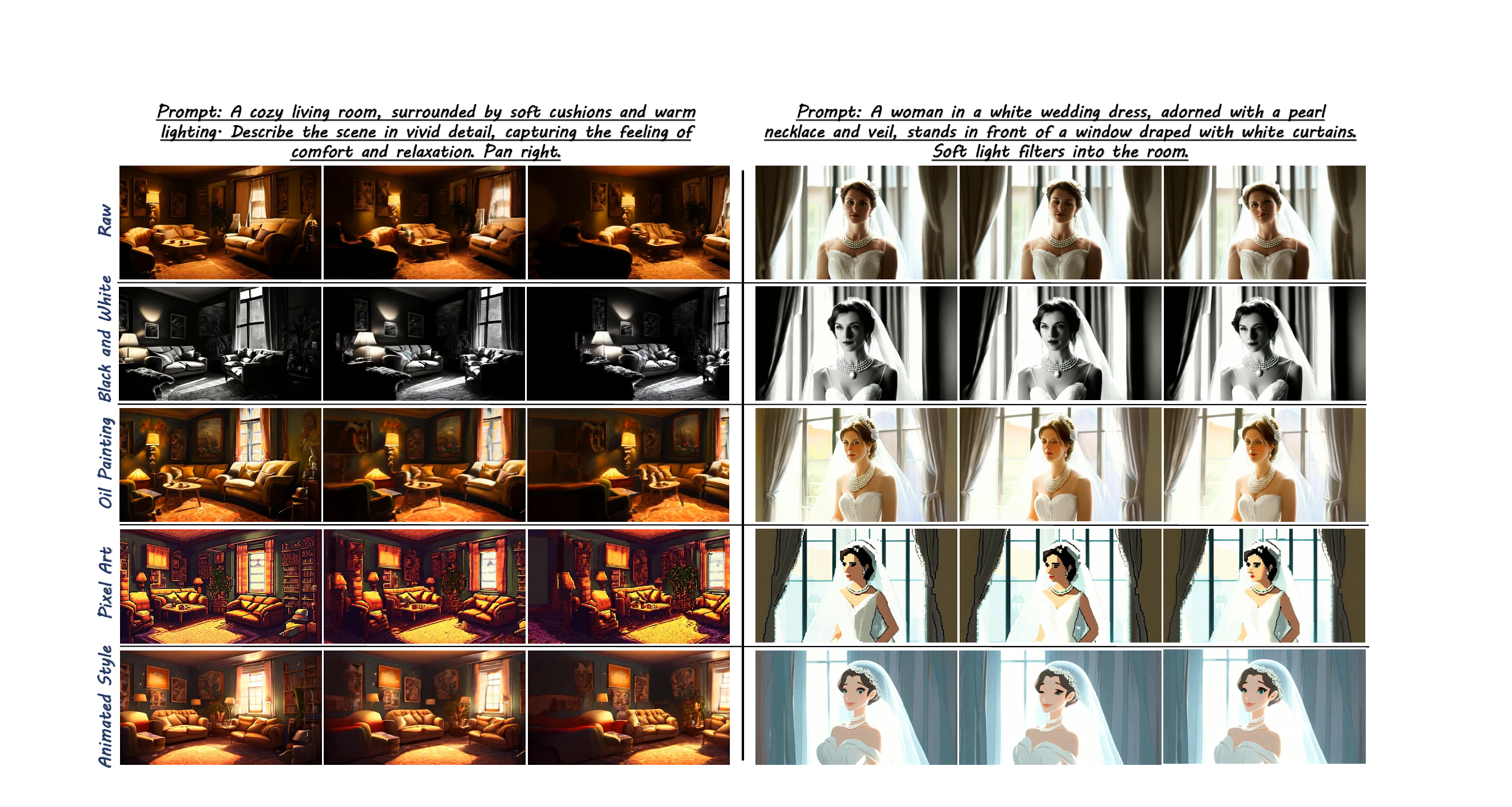}
    \caption{Two examples of combining consistency UFO and stylization UFO. In these cases, all the stylization UFOs have \( \alpha=1 \) and the Consistency UFOs have \( \alpha=0.1 \), all generated by OpenSoraV1.2 with a resolution of $720 \times 1280$ and a duration of 4 seconds.}
\end{figure*}

\textbf{Visualization of Consistency UFO with Different \( \alpha \).} An intuitive comparison, as shown in Figure 11, indicates that regardless of which model is used, in cases where the inconsistency in the generated videos is severe, using a low-intensity consistency UFO (\( \alpha \leq 0.2 \)) can significantly improve the inconsistencies in the produced videos. However, when \( \alpha > 0.2 \), although consistency is enhanced, the motion in the videos begins to decrease, yet the videos still retain some degree of motion and preserve the original elements, as seen in the example with \( \alpha = 0.3 \) in the figure. As \( \alpha \) continues to increase, such as at \( \alpha = 0.5 \), the generated videos largely lose their motion but still maintain the basic elements of the original image. When \( \alpha = 1 \), a completely static video that corresponds to the text is obtained, but the content and style of the image may significantly change compared to the videos generated by the original model.

\begin{figure*}[!t]
    \centering
    \includegraphics[width=1\linewidth]{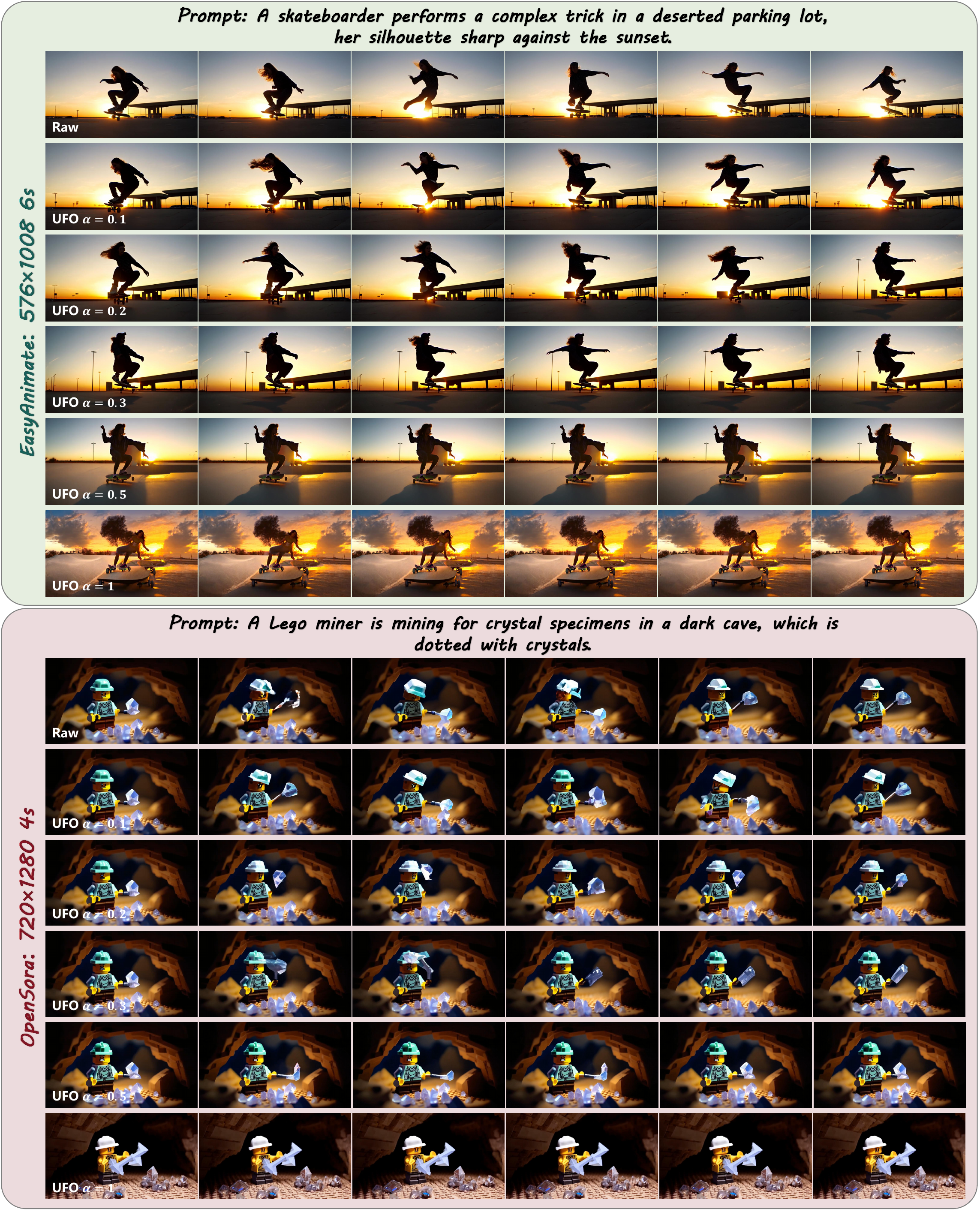}
    \caption{Example videos generated using consistency UFO with different \( \alpha \) values.}
\end{figure*}

\textbf{More Cases for Consistency UFO.} Figure 12 supplements cases where the consistency UFO is used across different models. In these examples, by comparing the changes in the video subjects before and after applying the UFO, one can clearly perceive an enhancement in video consistency.

\begin{figure*}[!t]
    \centering
    \includegraphics[width=1\linewidth]{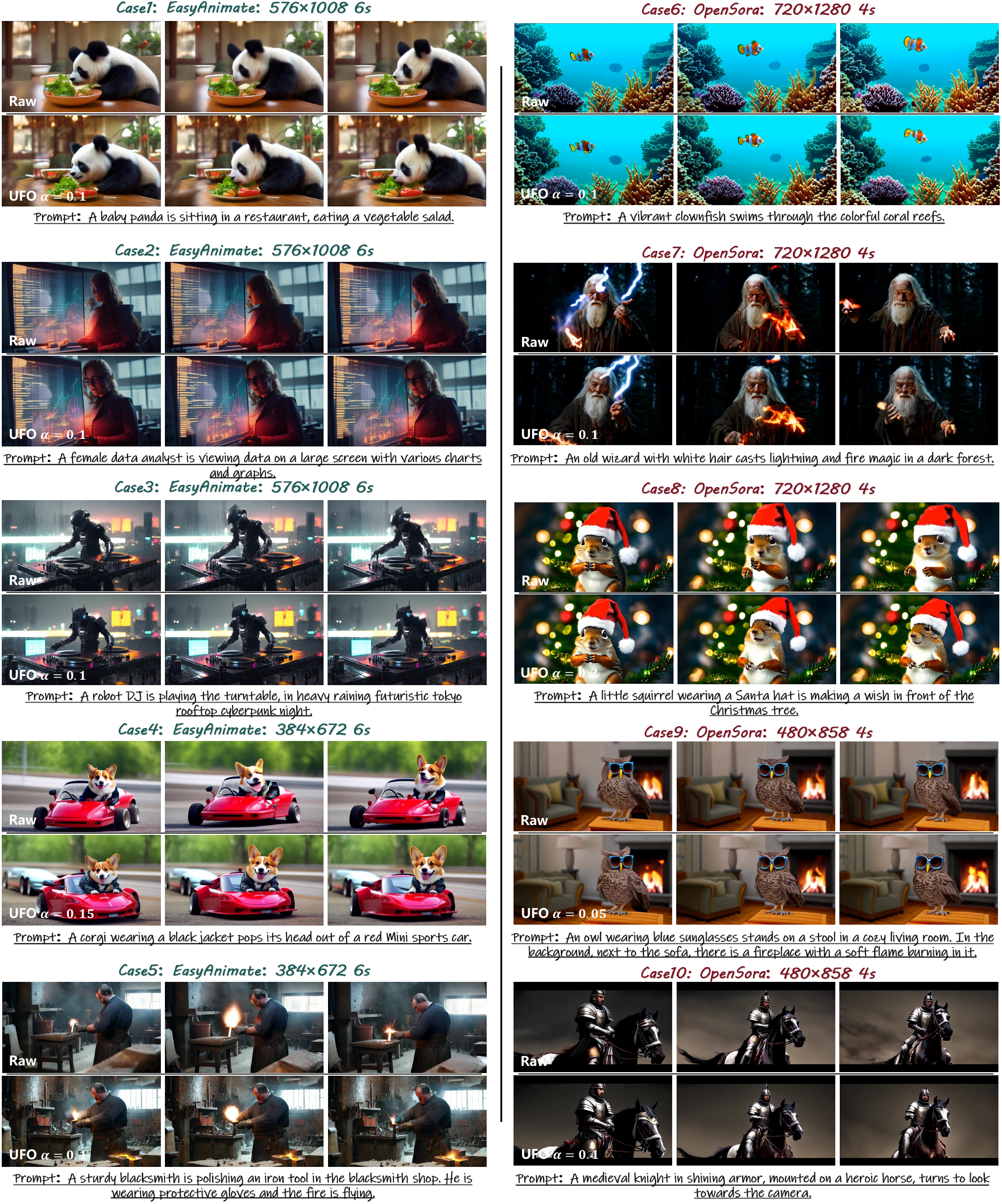}
    \caption{Examples comparing the intuitive effects of using the consistency UFO across different models and resolutions.}
\end{figure*}

\end{document}